\documentclass[letterpaper, 10 pt, conference]{ieeeconf}  

\IEEEoverridecommandlockouts                              

\usepackage{graphics} 
\usepackage{epsfig} 
\usepackage{times} 

\usepackage{bm}
\usepackage{color}

\usepackage{graphicx}
\usepackage{amsmath}
\usepackage{amsfonts}
\usepackage{amssymb}
\usepackage{url}
\usepackage{multirow}
\usepackage{algorithm}
\usepackage{algorithmic}

\usepackage{psfrag}
\usepackage{indentfirst}
\usepackage{subfigure} 

\usepackage{cite}
\usepackage{comment}

\usepackage{arydshln}
\usepackage{booktabs}
\usepackage{array}
\usepackage{makecell}
\usepackage{multirow}
\usepackage{tabu}
\usepackage{diagbox}

\providecommand{\keywords}[1]{\textbf{\textit{Index terms---}} #1}

\usepackage{geometry}
\title{\vspace{6.3mm} \LARGE  \bf
Multi-Agent Embodied Visual Semantic Navigation with Scene Prior Knowledge
}

\author{Xinzhu Liu, Di Guo, Huaping Liu$^{*}$, and Fuchun Sun
\thanks{The authors are with the Department of Computer Science and Technology, Tsinghua University, Beijing 100084, China, and also with the State Key Laboratory of Intelligent Technology and Systems, Beijing National Research Center for Information Science and Technology, Tsinghua University, Beijing 100084, China.}
\thanks{$^{*}$ Corresponding author: Huaping Liu (e-mail: hpliu@tsinghua.edu.cn)}%
}

\begin{document}

\maketitle
\thispagestyle{empty}
\pagestyle{empty}


\begin{abstract}
	

In visual semantic navigation, the robot navigates to a target object with egocentric visual observations and the class label of the target is given. It is a meaningful task inspiring a surge of relevant research. However, most of the existing models are only effective for single-agent navigation, and a single agent has low efficiency and poor fault tolerance when completing more complicated tasks. Multi-agent collaboration can improve the efficiency and has strong application potentials. In this paper, we propose the multi-agent visual semantic navigation, in which multiple agents collaborate with others to find multiple target objects. It is a challenging task that requires agents to learn reasonable collaboration strategies to perform efficient exploration under the restrictions of communication bandwidth. We develop a hierarchical decision framework based on semantic mapping, scene prior knowledge, and communication mechanism to solve this task. The results of testing experiments in unseen scenes with both known objects and unknown objects illustrate the higher accuracy and efficiency of the proposed model compared with the single-agent model.
	
\end{abstract}

\section{Introduction}

Visual semantic navigation is a task in which the agent is required to navigate to a target object utilizing the egocentric visual observations to perform actions to interact with the environment, given the class label of the target. It is of vital importance in the field of intelligent robots and has wide application prospects including home service robots for family use, warehouse logistics robots in industrial transportation, rescue robots in dangerous environments, and many other scenarios in the real world. In addition to its direct practical applications, visual semantic navigation is also beneficial for other challenging embodied tasks, such as embodied question answering \cite{das2018embodied}, visual-and-language navigation \cite{anderson2018vision}, sounding object navigation \cite{gan2020look} and so on.

Visual semantic navigation has inspired a large number of researchers to tackle this meaningful task. Learning-based frameworks have been widely used in this task. Reinforcement learning models including A3C \cite{druon2020visual, moghaddam2021learning}, DQN \cite{liang2020sscnav}, PPO \cite{campari2020exploiting, chaplot2020object}, and hierarchical reinforcement learning \cite{ye2021efficient} are exploited as the decision module for the agent to perform actions in the navigation task. Methodologies based on imitation learning \cite{wu2020towards}, supervised learning \cite{mousavian2019visual}, and unsupervised learning \cite{li2020unsupervised} have been proposed to learn the optimal navigation strategy. To improve the generalization ability of the decision model in unseen scenes, a self-adaptive meta-learning approach has been developed \cite{wortsman2019learning}. Besides, the attention mechanism \cite{mayo2021visual}, the memory unit \cite{de2020deep}, and the information-theoretic regularization\cite{wu2021reinforcement} have also been incorporated into the visual navigation model. However, most of the proposed models mentioned above are only effective in the single-agent setting where a single agent accomplishes a single task at a time. When humans demand the robot search for several different objects in a room, a single robot would have low efficiency in this mission because it can only search for all the target objects in succession. Meanwhile, the single agent has poor fault tolerance. If the judgment of the agent is wrong at a certain step, it may directly lead to the failure of the entire task.

\begin{figure}
	\centering
	\includegraphics[width=3.3in]{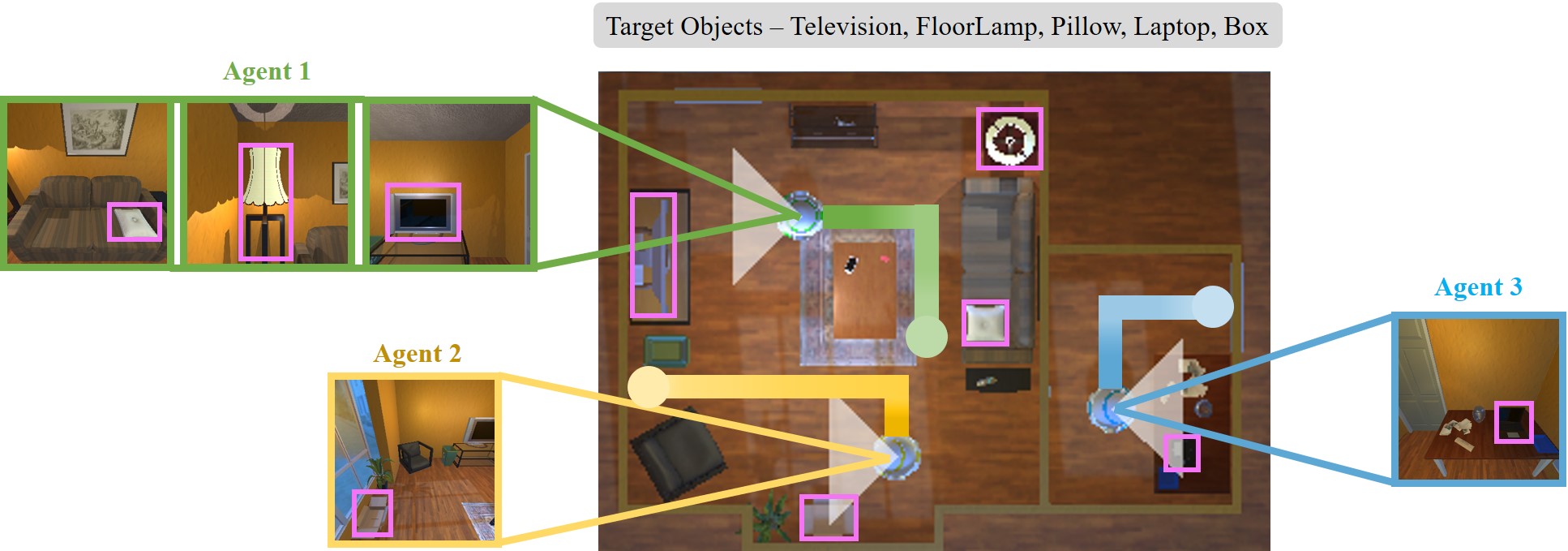}
	\caption{The multi-agent embodied visual semantic navigation task. In this scene, there are three collaborative agents aiming to find five target objects. The target objects are marked with pink boxes in the top-down map. Paths of agents are represented by rectangles with the gradient color, starting from circles. These three agents need to cooperate with each other to find all target objects with egocentric visual observations.}
	\label{fig:scene}
\end{figure}

Collaboration among multiple different robots has important research significance, and researchers hope that robots can collaborate with other robots to accomplish complex missions that are tough and inefficient for a single robot to complete. Multi-agent collaboration would have meaningful practical applications in industrial manufacturing, military cooperation, family service, and a great number of other fields in the future. In the early research, researchers have mainly studied the simple disembodied task for the multi-agent setting \cite{omidshafiei2017deep, gupta2017cooperative}. Recently, multi-agent decision models have been proposed to solve several multi-agent embodied tasks such as hide-and-seek task \cite{chen2019visual}, hiding game \cite{weihs2019artificial} and the Quake III game\cite{jaderberg2019human}. Communication mechanism has also been studied \cite{das2019tarmac} and applicated in a navigation task in which all agents need to find one same target through communication. \emph{FurnLift} \cite{jain2019two} and \emph{FurnMove} \cite{jain2020cordial} tasks are newly proposed to study multi-agent cooperation in embodied tasks. However, the proposed approaches for \emph{FurnLift} and \emph{FurnMove} mainly tackle the circumstances that two agents work together to complete a single task. It has not considered the situation in which more agents solve complex tasks collaboratively with multiple subtasks.


Furthermore, when the agent explores the environment in the visual semantic navigation task, the scene prior knowledge, which extracts the common spatial relationships between different objects, can effectively assist the agent in moving towards the target object with fewer steps. The scene prior knowledge has drawn great attention of researchers, and it is utilized to build the spatial and semantic relationships \cite{yang2018visual, nguyen2019reinforcement, qiu2020learning} in visual semantic navigation. It is also beneficial for multi-agent embodied tasks.

To study the collaboration among multiple agents in more complicated tasks, we propose the multi-agent visual semantic navigation task, in which multiple agents collaborate with others to find multiple different objects with the egocentric visual observations and the class labels of target objects, as is shown in Fig. \ref{fig:scene}. Multiple agents search for multiple targets simultaneously, which can improve the efficiency compared with a single agent, and the information obtained by different agents can be shared with others to avoid the failure of the entire task due to the decision mistake of one agent. On the other hand, multi-agent visual semantic navigation is a challenging task. First, it is difficult for multiple agents to learn effective strategies to avoid invalid exploration and improve the efficiency while ensuring the success rate of the navigation task. Second, there exists a bandwidth restriction in communication among multiple agents.



In this paper, we leverage the advantages of multi-agent exploration and scene prior knowledge to solve the embodied visual semantic navigation problem. The main contributions of this work are summarized as follows:

\begin{enumerate}
	\item A novel task of multi-agent embodied visual semantic navigation is proposed, and we extend the single-agent embodied visual semantic navigation to the multi-agent setting.
	
	\item We develop a hierarchical decision framework for multiple collaborative agents with the semantic map, scene prior knowledge, and communication mechanism to solve the multi-agent visual semantic navigation task.
	
	\item Experiments conducted on AI2-THOR\cite{ai2thor} and RoboTHOR\cite{robothor} demonstrate that the proposed multi-agent decision strategy achieves higher accuracy and efficiency compared with a single agent in visual semantic navigation. 
	
\end{enumerate}

\begin{figure*}
	\centering
	\includegraphics[width=6.9in]{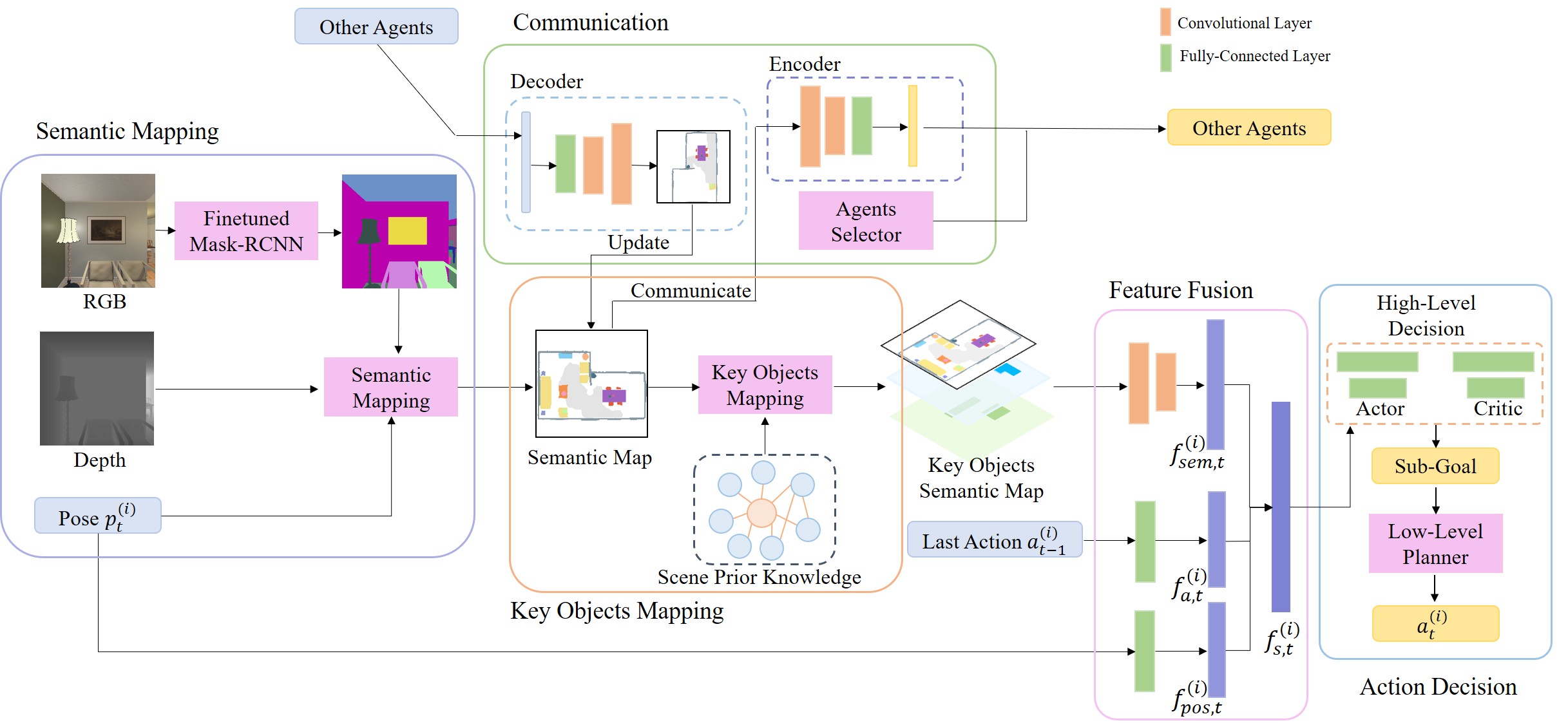}
	\caption{The structure of the proposed model for multi-agent visual semantic navigation. This model consists of five main modules: the semantic mapping module, the key objects mapping module, the communication module, the feature fusion module, and the action decision module.}
	\label{fig:model}
\end{figure*}

\section{Problem Formulation}

In the multi-agent embodied visual semantic navigation task, multiple agents aim to collaboratively find multiple objects with egocentric visual observations and the class labels of target objects. 

In a scene $\mathcal{S}$, there exist multiple objects of different categories which are denoted as $\mathcal{O}=\{{O}^{(1)}, {O}^{(2)},\cdots,{O}^{(K)}\}$, where each ${O}^{(k)}(k = 1, 2, \cdots, K)$ represents a certain type of the objects, and $K$ is the number of existing object types in $\mathcal{S}$. There are $N$ collaborative agents ${A}^{(1)}, {A}^{(2)},\cdots,{A}^{(N)}$. The target objects in the task is denoted as the set $\mathcal{T}=\{{T}^{(1)}, {T}^{(2)},\cdots,{T}^{(M)}\}$, in which $M$ represents the number of target object categories, and satisfies $\mathcal{T} \subseteq \mathcal{O}$. The goal of these collaborative agents is to find objects from all the categories in $\mathcal{T}$, and if there are multiple instances of a certain object type in the scene, agents only needs to find one of them.

At every timestep $t$, the $i$-th agent ${A}^{(i)}$ can obtain the egocentric visual observation $v_t^{(i)}$, the communication information from other agents $c_t^{(i)}$, the history state information $h_{t-1}^{(i)}$ as well as the class labels of targets $\mathcal{T}$. Every agent $A^{(i)}$ learns a proper action strategy ${\pi}^{(i)}$ to take action $a_t^{(i)}$ at timestep $t$, where $ a_t^{(i)} = {\pi}^{(i)}(v_t^{(i)}, c_t^{(i)}, h_{t-1}^{(i)}, \mathcal{T}) $. The action $a_t^{(i)}$ is chosen from the action set $\mathcal{A}=\{MoveAhead,\; \allowbreak
RotateRight,\; \allowbreak
RotateLeft,\; \allowbreak
LookUp,\; \allowbreak
LookDown,\; \allowbreak
Found,\; \allowbreak
Done\}. $
After every agent ${A}^{(i)}$ executes their action $a_t^{(i)}$, they obtain the new state of the environment including visual observation $v_{t+1}^{(i)}$, communication information $c_{t+1}^{(i)}$, and history information $h_{t}^{(i)}$. Agents then continue deciding actions with ${\pi}^{(i)}$ in the above manner until all agents peform $Done$ or the maximum number of steps is reached. The objective of this task is to learn optimal decision policy ${\pi}^{(i)}$ for every agent ${A}^{(i)}$ to collaborately find all target objects with as few steps as possible.

\section{Model Architecture}

The proposed model is composed of five main modules: the semantic mapping module, the key objects mapping module, the communication module, the feature fusion module, and the action decision module, which is demonstrated in Fig. \ref{fig:model}. The semantic mapping module generates a top-down semantic map with RGB and depth images of egocentric visual observations. The key objects mapping module marks the location of key objects with the scene prior knowledge. The communication module is responsible for transmitting and receiving information from other agents. The feature fusion module extracts the feature of the target objects and the history states. Then the action decision module decides the next action of the agent to perform.

\subsection{Semantic Mapping}

The semantic mapping module utilizes the pre-trained Mask-RCNN \cite{he2017mask} to obtain the semantic segmentation results of the RGB images. Then the semantic segmentation results, the depth image, and the pose of the agent are used to project each pixel with a specific object category into the 3D world coordinate to obtain a 3D voxel semantic map. We sum along the height dimension of the voxel semantic map to get a top-down 2D semantic map of size $L \times W \times (K_{total} + 2)$, in which the length and width of the map are $L$ and $W$ respectively. Each element in the map denotes a grid of size $5cm \times 5cm$ in simulation scenes. $K_{total}$ is the number of the object categories, and the first $K_{total}$ channels of the semantic map denote the area of the corresponding object categories. The last two channels represent the occupied and explored area respectively.



To generate the semantic map more accurately, the pre-trained Mask-RCNN is finetuned in training scenes. We have obtained the egocentric RGB images and the true segmentation results from the AI2-THOR and RoboTHOR platforms. Then we finetune the Mask-RCNN model in generated images to obtain a more suitable model for the object categories in experiment scenes. We utilize the finetuned segmentation model in the semantic mapping process.

\subsection{Key Objects Mapping}

The key objects mapping module utilizes the scene prior knowledge to find the key objects with the current visual observation and generates the key objects map of size $L \times W \times 2$ based on the 2D semantic map. The key objects include the target objects and the objects which have a common spatial relationship with target objects. 


\subsubsection{Scene Prior Knowledge}

\begin{figure}
	\centering
	\includegraphics[width=2.1in]{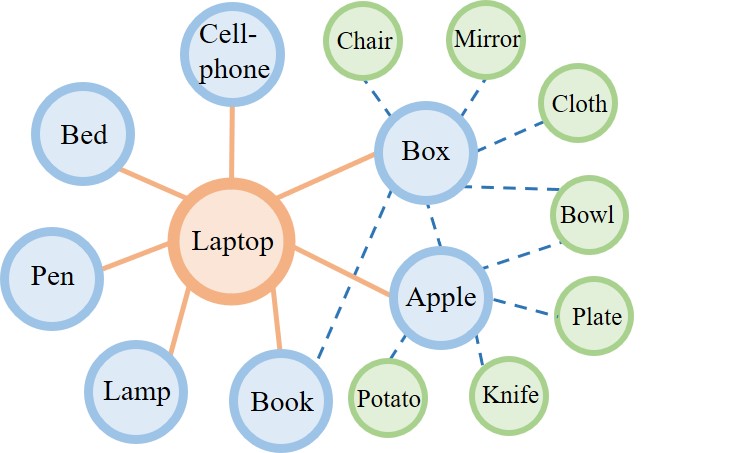}
	\caption{A part of the scene prior knowledge. The solid lines represent the relationships between Laptop and other objects. The dotted lines denote the relationships between Box, Apple, and other objects. There exist other edges between each node in this figure and other undrawn nodes which denote the relationships between different objects.}
	\label{fig:sceneprior}
\end{figure}

We build the scene prior knowledge graph using the method similar to Ref. \cite{yang2018visual}. The scene prior knowledge graph is represented as $\mathcal{G}=(Q,E)$, in which each node $q \in Q$ denotes a class of object, and each edge $e \in E$ represents a specific relationship between two object categories, and a part of the built graph is shown in Fig. \ref{fig:sceneprior}. In this task, we extract the object spatial relationships for all object categories in AI2-THOR and RoboTHOR from the Visual Genome \cite{krishna2017visual} to form the scene prior knowledge graph.

\subsubsection{Key Objects Extraction}

The key objects map is constructed by marking the key objects based on the semantic map. There are two layers in the key objects map, and the size of each layer $L \times W$ is the same as the semantic map. The first layer marks the target objects and the second layer marks the other key objects which have spatial relationships with target objects. This module judges whether there exist key objects in the current visual observation according to the semantic segmentation results. That is, the module judges if there are target objects and objects that have relationships with target objects using the scene prior knowledge graph. If there do not exist any key objects, the key objects map would be denoted as a matrix with all zero elements. If there exist key objects within the view, the module would convert the location of the key objects in visual images to the location in a 2D top-down map with the method similar to the semantic mapping process, and then mark the corresponding pixels in the map. Meanwhile, the model would mark all key object instances of the same categories based on the built semantic map. 
These marked key objects can help agents move to the position as close to target objects as possible.



\subsection{Communication}

The communication module includes two parts: the information processing part and the agent selection part. The information processing part is utilized to encode and decode the relevant information. The agent selection part chooses which agent the encoded information is transmitted to according to the specific regulation.

\subsubsection{Information Processing}

In order to decrease the amount of data transmission during communication and reduce the required communication bandwidth, the information processing part, composed of an encoder and a decoder, is utilized to encode the semantic information into a vector to reduce the amount of communication information. The encoder consists of two convolutional layers as well as a fully-connected layer and can encode the constructed semantic map into a one-dimensional vector. The decoder is composed of a fully-connected layer as well as two upsample convolutional layers, and is utilized to decode the received communication vector into a semantic matrix. We train the encoder and decoder in advance. The original semantic maps are input into the encoder and the decoder successively to generate recovered semantic maps, and we use the original semantic maps as supervision to train the encoder and decoder. The difference between the original and the recovered semantic maps should be as small as possible. 

\subsubsection{Agent Selection}

The agent selection part chooses agents to whom the communication information is sent before each round of communication. Every agent selects the agents to communicate who are more likely to effectively utilize the communication information, which saves communication bandwidth compared with directly broadcasting to all agents. In this task, the communication area of each agent is denoted as the area where each side of its field of view extends 1.0 meter outwards. In each round of communication, agents only send the encoded information to other agents in their communication area, as agents in the communication area are much likely to use the semantic map information in the next few steps.

\subsection{Feature Fusion}

The feature fusion module takes the built semantic map, the key objects map, the current pose, and the last performed action as input to extract the feature of the current state of the agent. At timestep $t$, the semantic map and key objects map are concatenated into a top-down key objects semantic map of size $L \times W \times (K_{total} + 4)$. We use the CNN network with two convolutional layers to generate the semantic feature $f_{sem,t}^{(i)}$ of the map preserving the spatial information. A fully-connected layer is utilized to obtain the pose feature $f_{pos,t}^{(i)}$ from the current pose $p_{t}^{(i)}$, mapping the pose feature to the same feature space of the semantic feature. Meanwhile, another fully-connected layer is used to project the last action vector $a_{t-1}^{(i)}$ to the same feature space generating the action feature $f_{a,t}^{(i)}$. Then, we concatenate feature $f_{sem,t}^{(i)}$, $f_{pos,t}^{(i)}$, and $f_{a,t}^{(i)}$ to get the integral state feature $f_{s,t}^{(i)}$.

\subsection{Action Decision}

The action decision module takes the state feature $f_{s,t}^{(i)}$ and decides the next action $a_t^{(i)}$ to perform for the agent $A^{(i)}$ at timestep $t$. The action decision module is a hierarchical decision module, which is composed of a high-level decision part and a low-level decision part. 

\subsubsection{High-Level Decision}

The high-level decision part decides a sub-goal $sg_t^{(i)}=(sgx_t^{(i)}, sgy_t^{(i)})$ on the top-down map for the agent $A^{(i)}$ based on the state feature $f_{s,t}^{(i)}$. The decision part is trained with PPO \cite{schulman2017proximal} reinforcement learning algorithm, which includes an actor and a critic composed of fully-connected layers. It generates the sub-goal every $d$ steps (in our experiments $d=10$). If the interval steps $d$ are reached or the agent reaches the sub-goal with the low-level decision part within $d$ steps, the high-level decision part would re-plan the sub-goal. The reward of this strategy is,
\begin{align}
	& R_{u}= \alpha \Delta dis_{to}(sg_u, sg_{u-1}) + \beta \Delta dis_{ko}(sg_u, sg_{u-1})       \nonumber \\
	&\Delta dis_{to}(sg_u, sg_{u-1}) = dis_{to}(sg_{u-1}) - dis_{to}(sg_u)  \label{eq2} \\
	&\Delta dis_{ko}(sg_u, sg_{u-1}) = dis_{ko}(sg_{u-1}) - dis_{ko}(sg_u)\nonumber
\end{align}
where $dis_{to}(sg_u)$ represents the distance between the sub-goal $sg_u$ and the nearest target object, and $dis_{ko}(sg_u)$ denotes the distance between $sg_u$ and the nearest key objects except for target objects. $\Delta dis_{to}(sg_u, sg_{u-1})$ is the decrease in distance between the new sub-goal and the nearest target object, and if the value is greater than 0, the new sub-goal is closer to the target object compared with the old one. The same is true for $\Delta dis_{ko}$. $\alpha$ and $\beta$ are hyperparameters to control the calculation of the reward function, and we set $\alpha = 0.7, \beta = 0.3$ in this task. This reward function denotes that both the distance between the sub-goal and target objects as well as the distance between the sub-goal and key objects are taken into consideration, incorporating the effect of the scene prior knowledge.

\subsubsection{Low-Level Decision}

The low-level decision part decides the specific actions for the agent using the modified breadth-first search \cite{bundy1984breadth} shortest path algorithm to reach the sub-goal generated by the high-level decision part. The low-level decision module generates the action based on the updated semantic map every step, in which the occupied channel denotes the obstacle position and the explored channel represents the region that has been walked through.

\renewcommand{\arraystretch}{1.1}
\begin{table*}[!t] \caption{Quantitative Results in Multi-agent Embodied Visual Semantic Navigation in AI2-THOR}
	\label{table1}
	\centering
	\setlength{\tabcolsep}{0.55mm}
	\begin{tabular}{c|c|ccc|ccc|ccc|ccc|ccc}
		\toprule[1pt]
		\multirow{2}{*}{Testing Setting} & \multirow{2}{*}{Methods} 
		& \multicolumn{3}{c|}{$N=1$} & \multicolumn{3}{c|}{$N=2$} & \multicolumn{3}{c|}{$N=3$} & \multicolumn{3}{c|}{$N=4$} & \multicolumn{3}{c}{$N=5$}         \\ 
		& & SR(\%)  & SPL(\%) & EI(\%)  & SR(\%) & SPL(\%) & EI(\%)  & SR(\%) & SPL(\%) & EI(\%)   & SR(\%)  & SPL(\%) & EI(\%)  & SR(\%) & SPL(\%) & EI(\%) \\ 
		
		\midrule [0.8pt] 
		
		\multirow{7}{*}{\begin{tabular}[c]{@{}c@{}}Unseen Scenes\\ Known Objects\end{tabular}} & Random  & 8.02  & 3.12  & -  & 8.31  & 3.22  & 10.01  & 8.33  & 3.35 & 12.10       & 8.51  & 3.76 & 12.25 & 8.55 & 3.79 & 12.27 \\ 
		
		& RGB+IR  & 23.30  &  13.51  & -  &  23.49  & 13.60 & 25.50  & 24.05  & 13.69 & 35.15   & 24.39  & 14.06 & 35.69 & 24.45 & 14.30 & 37.01 \\ 
		
		& RGB+SP.+IR & 25.63  &  14.77  & -  &  26.13  & 14.86 & 28.33  & 27.29  & 15.06 & 37.23   & 28.17  & 16.30 & 37.36 & 28.35 & 16.43 & 37.69 \\
		
		& Cordial Sync \cite{jain2020cordial}  & 35.73  & 14.88  & -    & 36.25 &  15.30      &  28.56 &  36.80  & 16.61  & 38.81  & 37.02   & 17.03 & 39.23 & 37.15 & 17.82 & 39.35 \\
		
		& Central w/o SP.   & 41.00  & 15.10 & - & 43.17  & 15.56 & 29.07 & 44.03  & 18.01 & 40.03 & 44.28   & 18.32 & 40.31  & 44.50   & 18.39 & 40.36  \\ 
		
		& \textbf{Ours}  & \textbf{45.19}    & \textbf{18.25} & -  & \textbf{47.02}  & \textbf{19.13} & \textbf{31.01} & \textbf{47.60}  & \textbf{21.99} & \textbf{41.33} & \textbf{47.77}  & \textbf{23.25} & \textbf{43.55}  & \textbf{48.02}   & \textbf{23.30} & \textbf{43.60}    \\ 
		\cdashline{2-17}[1.5pt/1.5pt]
		
		& \emph{Central}$^{\star}$  & \emph{45.19}    & \emph{18.25} &-  &\emph{47.93}   &\emph{19.66}  &\emph{32.10}  &\emph{48.16}   &\emph{22.33}  & \emph{42.00} & \emph{48.71}  &\emph{23.56}   &\emph{43.59}  & \emph{48.90}   &\emph{23.69}  & \emph{43.75}   \\ 
		

		\midrule [0.8pt]
		
		\multirow{7}{*}{\begin{tabular}[c]{@{}c@{}}Unseen Scenes\\ Unknown Objects\end{tabular}} & Random  & 8.00  & 3.13  & -  & 8.28  & 3.20  & 9.96  & 8.31  & 3.32 & 12.05     & 8.45  & 3.76 & 12.09 & 8.53 & 3.70 & 12.12 \\ 
		
		& RGB+IR  & 16.05  &  6.89  & -  &  16.33  & 7.02 & 20.03  & 17.27  & 7.55 & 30.76   & 17.66  & 7.90 & 33.77 & 17.70 & 8.02 & 33.90 \\ 
		
		& RGB+SP.+IR & 20.25  &  9.95  & -  &  21.33  & 10.55 & 25.74  & 22.05  & 11.12 & 33.32   & 23.35  & 12.03 & 35.66 & 23.58 & 12.25 & 36.05 \\
		
		& Cordial Sync \cite{jain2020cordial}  & 32.55  & 12.03  & -    & 33.19 &  12.88      &  27.35 &  34.12  & 13.27  & 35.20  & 35.33   & 14.85 & 36.55 & 35.63 & 15.05 & 36.61 \\
		
		& Central w/o SP.   & 39.73  & 13.06 & - & 40.20  & 13.75 & 28.35 & 41.88  & 15.12 & 39.12 & 43.01   & 17.00 & 40.17  & 43.29   & 17.22 & 40.26  \\ 
		
		& \textbf{Ours}  & \textbf{43.81}    & \textbf{16.35} & -  & \textbf{43.95}  & \textbf{16.55} & \textbf{30.00} & \textbf{44.56}  & \textbf{17.25} & \textbf{39.05} & \textbf{45.25}  & \textbf{17.93} & \textbf{39.98}  & \textbf{46.17}   & \textbf{18.07} & \textbf{40.02}    \\
		\cdashline{2-17}[1.5pt/1.5pt]
		
		& \emph{Central}$^{\star}$  & \emph{43.81}    & \emph{16.35} &-  &\emph{43.98}   &\emph{17.01}  &\emph{30.75}  &\emph{45.15}   &\emph{18.85}  & \emph{40.33} & \emph{46.33}  &\emph{19.60}   &\emph{41.35}  & \emph{46.72}   &\emph{19.77}  & \emph{41.66}   \\ 
		

		\bottomrule[1pt]
	\end{tabular}
\end{table*}

\renewcommand{\arraystretch}{1.1}
\begin{table*}[!t] \caption{Quantitative Results in Multi-agent Embodied Visual Semantic Navigation in RoboTHOR}
	\label{table2}
	\centering
	\setlength{\tabcolsep}{0.55mm}
	\begin{tabular}{c|c|ccc|ccc|ccc|ccc|ccc}
		\toprule[1pt]
		\multirow{2}{*}{Testing Setting} & \multirow{2}{*}{Methods}   & \multicolumn{3}{c|}{$N=1$} & \multicolumn{3}{c|}{$N=2$} & \multicolumn{3}{c|}{$N=3$} & \multicolumn{3}{c|}{$N=4$} & \multicolumn{3}{c}{$N=5$}         \\ 
		& & SR(\%)  & SPL(\%) & EI(\%)  & SR(\%) & SPL(\%) & EI(\%)  & SR(\%) & SPL(\%) & EI(\%)   & SR(\%)  & SPL(\%) & EI(\%)  & SR(\%) & SPL(\%) & EI(\%) \\
		
		\midrule [0.7pt] 
		
		\multirow{7}{*}{\begin{tabular}[c]{@{}c@{}}Unseen Scenes\\ Known Objects\end{tabular}} & Random  & 6.75  & 2.83  & -  & 6.83  & 2.91  & 9.35  & 6.88  & 3.01 & 11.35 & 7.12 & 3.09 & 11.39 & 7.30 & 3.55 & 11.88 \\
		
		& RGB+IR  & 19.72  & 10.33  & -  &  20.03  & 10.52 & 23.15  & 21.85  & 11.08 & 33.90   & 22.12  & 12.01 & 34.15 & 22.39 & 12.25 & 34.23 \\ 
		
		& RGB+SP.+IR & 20.90  & 11.05 & - & 21.76  & 11.69 & 25.70  & 23.15  & 12.76 & 35.01   & 24.09  & 12.90 & 35.80 & 24.87 & 13.15 & 35.88 \\
		
		& Cordial Sync \cite{jain2020cordial}  & 31.87  &  11.77  & -  &  32.22  & 12.30       &  26.65  & 34.05   &  13.65   &  38.02  &  35.11  & 14.63 & 39.01 & 35.65 & 14.88 & 39.20 \\

		& Central w/o SP.  & 36.15  & 12.60 & - & 37.76  & 12.99 & 28.88 & 38.95  & 15.03 & 39.95 & 39.16   & 15.25 & 40.35  & 39.17   & 15.30 & 40.39  \\ 
		
		& \textbf{Ours}      & \textbf{40.85}    & \textbf{13.95} & -  & \textbf{42.20}  & \textbf{14.17} & \textbf{30.55} & \textbf{42.33}  & \textbf{17.75} & \textbf{41.55} & \textbf{42.52}  & \textbf{19.06} & \textbf{43.62}  & \textbf{43.01} & \textbf{20.23} &  \textbf{43.65}   \\ 
		\cdashline{2-17}[1.5pt/1.5pt]

		& \emph{Central}$^{\star}$  & \emph{40.85}    &\emph{13.95}   & -   & \emph{42.76}   & \emph{15.08}  &\emph{31.63}    &  \emph{42.85}  &  \emph{18.05}  & \emph{42.05}  &  \emph{42.91}  & \emph{19.88}  & \emph{43.88}   & \emph{43.15}    &  \emph{20.33}  & \emph{43.93}  \\
		
		
		\midrule [0.7pt]
		
		\multirow{7}{*}{\begin{tabular}[c]{@{}c@{}}Unseen Scenes\\ Unknown Objects\end{tabular}} & Random  & 6.57  & 2.80  & -  & 6.83  & 2.88  & 9.25  & 6.85  & 2.97 & 11.30 & 7.10 & 3.00 & 11.36 & 7.22 & 3.32 & 11.55 \\
		
		& RGB+IR  & 13.36  & 5.89  & -  &  13.85  & 6.33 & 19.50  & 15.39  & 7.35 & 27.12 & 16.05  & 8.06 & 30.55 & 16.88 & 8.60 & 30.96 \\ 
		
		& RGB+SP.+IR & 15.39  & 8.25 & - & 16.48  & 9.13 & 22.12  & 17.32  & 10.01 & 32.50   & 18.85  & 11.58 & 33.23 & 19.05 & 11.33 & 33.70 \\
		
		& Cordial Sync \cite{jain2020cordial}  & 28.55 &  10.03   & -  &  29.12  &  11.31      &  24.85   &  31.56  &  12.05  &  36.77  &  33.35   & 12.90  & 37.50 & 33.62 & 13.05 & 37.89 \\

		& Central w/o SP.  & 33.49  & 11.88 & - & 34.88  & 12.05 & 26.10 & 35.23  & 13.15 & 37.53 & 36.00   & 14.07 & 39.42  & 36.23   & 14.57 & 39.55  \\ 
		
		& \textbf{Ours}      & \textbf{38.66}    & \textbf{12.12} & -  & \textbf{39.89}  & \textbf{12.65} & \textbf{29.36} & \textbf{40.88}  & \textbf{14.96} & \textbf{39.33} & \textbf{41.15}  & \textbf{15.50} & \textbf{40.85}  & \textbf{41.33} & \textbf{15.72} &  \textbf{40.99}   \\ \cdashline{2-17}[1.5pt/1.5pt]

		& \emph{Central}$^{\star}$  & \emph{38.66}    &\emph{12.12}   & -   & \emph{40.19}   & \emph{13.30}  &\emph{30.35}    &  \emph{41.33}  &  \emph{15.21}  & \emph{40.25}  &  \emph{41.89}  & \emph{16.33}  & \emph{41.25}   & \emph{42.07}    &  \emph{16.93}  & \emph{41.60}  \\ 
		
		
		\bottomrule[1pt]
	\end{tabular}
\end{table*}

\section{Experiment}

\subsection{Dataset}

We conduct experiments on two platform: AI2-THOR \cite{ai2thor} and RoboTHOR\cite{robothor}. AI2-THOR includes 120 indoor scenes from four room types: \emph{Kitchens}, \emph{Living rooms}, \emph{Bedrooms} and \emph{Bathrooms}. There are 80 training scenes and 40 testing scenes totally in AI2-THOR. RoboTHOR contains a more complicated room structure with more wall blocks, and there are a total of 60 training scenes and 15 testing scenes. Training scenes are different from testing scenes, and we evaluate our model in unseen testing scenes. In every scene, we select $M$ ($M=1,2,3,4,5$ respectively) target objects to form the navigation task and ensure that every target has at least one instance. In AI2-THOR, we generate 7900 training tasks and 4000 testing tasks. In RoboTHOR, we generate 5940 training tasks and 1485 testing tasks.


\begin{figure*}[!t]
	\centering
	\subfigure[Navigation process of two agents in AI2-THOR.]{
		\includegraphics[width=6.8in]{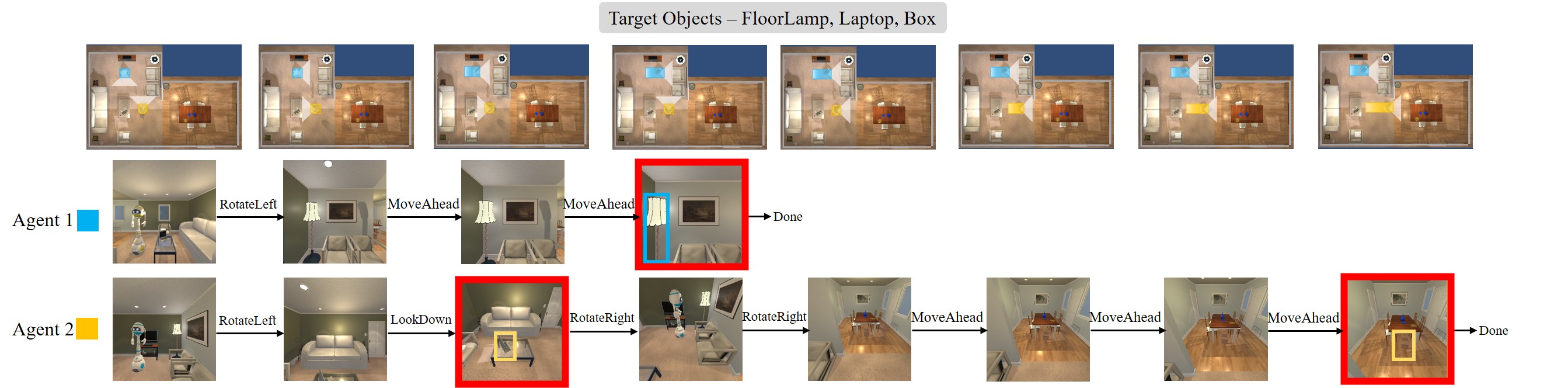}
	}
	\quad
	\subfigure[Navigation process of three agents in RoboTHOR.]{
		\includegraphics[width=6.8in]{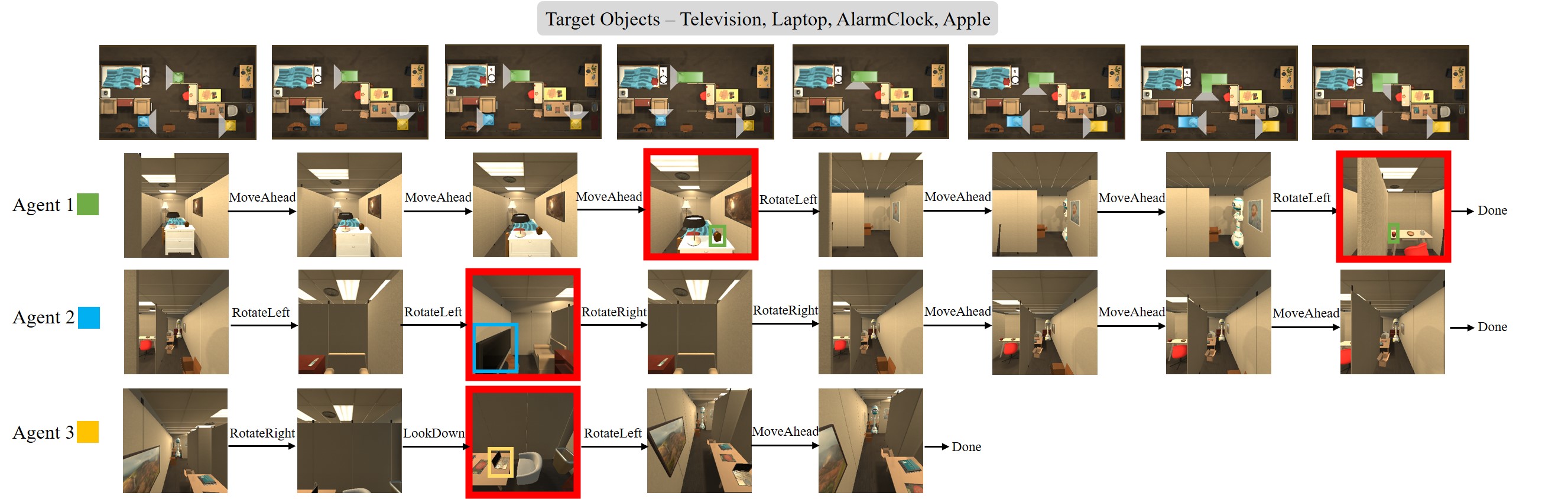}
	}
	\caption{Qualitative results of multi-agent visual semantic navigation. (a) shows the navigation process in \emph{FloorPlan223} in AI2-THOR, in which two agents search for three targets. Agent $1$ and Agent $2$ see Laptop and FloorLamp respectively initially. FloorLamp is closer to Agent $1$ and Laptop is closer to Agent $2$, but agents cannot see the target closer to them at first. They communicate with each other and navigate the object closer to them in a few steps of actions. Finally, Agent $1$ finds FloorLamp. Agent $2$ finds Laptop and continues to search for Box without stopping. (b) shows the navigation process in \emph{FloorPlan\_Val1\_4} in RoboTHOR, where three agents navigate to four targets. Agent $1$ navigates to AlarmClock and then continues to find Apple. Agent $2$ and Agent $3$ share the obtained information and navigate the object closer to them. Finally, Agent $2$ finds Television and Agent $3$ finds Laptop.
	}
	\label{fig:result}
\end{figure*}


\subsection{Implementation Details}

In our experiments, the number of target object is set to be $M=1,2,3,4,5$. 
In every task, $N$ agents are initialized at random reachable positions. We set $N=1,2,3,4,5$ respectively to perform multi-agent navigation in all tasks.

The action space is $\mathcal{A}=\{MoveAhead,\; \allowbreak
RotateRight,\; \allowbreak
RotateLeft,\; \allowbreak
LookUp,\; \allowbreak
LookDown,\; \allowbreak
Found, \; \allowbreak
Done\}$. $MoveAhead$ means moving straight along the orientation for $0.25m$. $RotateRight$ and $RotateLeft$ indicate turning right or left for 90 degrees respectively. $LookUp$ and $LookDown$ mean turning the view up or down for 30 degrees respectively. $Found$ means that the agent has found a target object. $Done$ means that the agent has finished all its actions and stops. When the agent performs $Found$, if the target object is within the view of the agent and the distance between the agent and the target is less than $1.0m$, then the target object is considered found successfully. When all agents perform $Done$, if all targets have been found successfully, the current task is successful, otherwise the task fails.

We train the encoder and decoder in several training scenes in advance. Twenty training scenes are chosen from the whole training set. We put an agent in every reachable pose in each scene to get the RGB and depth images, and generate 2D semantic maps $\{Sem_{1}, Sem_{2}, \cdots, Sem_{cn}\}$ with the method mentioned in section \uppercase\expandafter{\romannumeral4.A}, where $cn$ is the number of the generated semantic maps. Then we put the semantic maps into the encoder $\bm{En}$ and the decoder $\bm{De}$ successively to get the recovered maps results $Sre_i=\bm{De}(\bm{En}(Sem_i))$, where $i=1, 2, \cdots, cn$ and $Sre_i$ represents the recovered semantic maps. The loss function of this training process is defined as the pixel-wise distance between original semantic maps and recovered semantic maps,
\begin{align}
	Loss_{com}=\frac{1}{nc} \sum\limits_{i=1}^{nc} Dis(Sre_i, Sem_i), \label{eq1}
\end{align}
where $Dis$ denotes the pixel-wise distance between $Sre_i$ and $Sem_i$, and the distance should be as small as possible.

We train the high-level decision module with the PPO algorithm. We use Adam optimizer during training with a learning rate of 0.0001. We set the discount factor $\gamma = 0.99$.

\subsection{Baselines}
We compare our models with the following baselines to evaluate the effectiveness of the proposed model.

\subsubsection{Random}
$N$ agents randomly perform actions from the action space $\mathcal{A}$ at each timestep.

\subsubsection{RGB+RL}
Every agent takes egocentric RGB images and class labels of target objects as input and utilizes ResNet-18 \cite{he2016deep} to extract visual features. Every agent decides the action based on A3C \cite{mnih2016asynchronous} algorithm without communication with other agents. This baseline directly extends the general single-agent navigation model to the multi-agent setting.

\subsubsection{RGB+SP.+RL}
This model uses GCN to introduce the scene prior knowledge to \emph{RGB+RL}, and it extends the single-agent navigation model with scene priors \cite{yang2018visual} to multiple agents without communication.

\subsubsection{Central}
In this model, there is a central agent that has access to observations of all agents and makes decisions for agents based on the whole information. The other parts of this model are the same as our model.

\subsubsection{Central w/o SP.}
This model does not utilize the scene prior knowledge, and the rest is the same as \emph{Central}.

\subsubsection{Cordial Sync \cite{jain2020cordial}}
We utilize the structure of the decision policy proposed in \cite{jain2020cordial}, which is a decentralized model without scene prior knowledge, and modify it to apply to multi-agent navigation.

\renewcommand{\arraystretch}{1.05}
\begin{table*}[!t] \caption{Results of the Ablation Experiment in Multi-agent Visual Semantic Navigation in AI2-THOR}
	\label{table3}
	\centering
	\setlength{\tabcolsep}{0.55mm}
	\begin{tabular}{c|c|ccc|ccc|ccc|ccc|ccc}
		\toprule[1pt]
		\multirow{2}{*}{Testing Setting} & \multirow{2}{*}{Methods}   & \multicolumn{3}{c|}{$N=1$} & \multicolumn{3}{c|}{$N=2$} & \multicolumn{3}{c|}{$N=3$} & \multicolumn{3}{c|}{$N=4$} & \multicolumn{3}{c}{$N=5$}         \\ 
		& & SR(\%)  & SPL(\%) & EI(\%)  & SR(\%) & SPL(\%) & EI(\%)  & SR(\%) & SPL(\%) & EI(\%)   & SR(\%)  & SPL(\%) & EI(\%)  & SR(\%) & SPL(\%) & EI(\%) \\
		
		\midrule [0.8pt] 
		\multirow{4}{*}{\begin{tabular}[c]{@{}c@{}}Unseen Scenes\\ Known Objects\end{tabular}} & Ours w/o SemMap.  & 30.33  & 12.50  & -  & 31.85  & 13.39  & 23.33  & 33.65    & 15.06   &  25.68  & 34.09  & 15.25  & 27.30  & 34.21 & 15.30 & 27.52 \\
		
		& Ours w/o SP.  & 41.00  & 15.10  & -  & 42.85  & 15.21  & 28.55  & 43.25  & 17.15  & 39.02  & 43.31  & 18.01  & 39.95  & 43.50  & 18.23  & 40.01 \\ 
		
		& Ours w/o HieDec. & 38.88  &  14.76  &  -  &  40.33  & 15.09  & 28.08 & 42.87     & 17.03    & 39.01   & 43.22 & 17.75  & 39.66 & 43.39 & 18.00 & 39.63 \\ 
		
		& \textbf{Ours}   & \textbf{45.19}    & \textbf{18.25} & -  & \textbf{47.02}  & \textbf{19.13} & \textbf{31.01} & \textbf{47.60}  & \textbf{21.99}  & \textbf{41.33} & \textbf{47.77}  & \textbf{23.25} & \textbf{43.55}  & \textbf{48.02}   & \textbf{23.30} & \textbf{43.60} \\ 
		
		\midrule [0.8pt]
		
		\multirow{4}{*}{\begin{tabular}[c]{@{}c@{}}Unseen Scenes\\ Unknown Objects\end{tabular}} & Ours w/o SemMap.  & 23.50  & 7.21  & -  & 24.39  & 8.01  & 20.12  & 26.02    & 10.23   &  22.10  & 27.25  & 11.60  & 25.11  & 27.40 & 11.88 & 25.60 \\
		
		& Ours w/o SP.  & 39.73  & 13.06  & -  & 39.33  & 13.25  & 27.10  & 41.16  & 14.88  & 38.60  & 42.59  & 15.98  & 38.89  & 42.70  & 16.10  & 39.05 \\ 
		
		& Ours w/o HieDec. & 36.50  &  12.12  &  -  &  38.00  & 13.10  & 26.56 & 40.01     & 14.66    & 38.55   & 42.39 & 15.66  & 38.80 & 42.60 & 15.73 & 38.91 \\ 
		
		& \textbf{Ours}  & \textbf{43.81}    & \textbf{16.35} & -  & \textbf{43.95}  & \textbf{16.55} & \textbf{30.00} & \textbf{44.56}  & \textbf{17.25} & \textbf{39.05} & \textbf{45.25}  & \textbf{17.93} & \textbf{39.98}  & \textbf{46.17}   & \textbf{18.07} & \textbf{40.02}    \\
		
		\bottomrule[1pt]
	\end{tabular}
\end{table*}

\subsection{Evaluation Metrics}

We use three main metrics to evaluate the performance of our model: Success Rate (SR), Success weighted by Path Length (SPL) proposed in \cite{SPL} and Efficiency Improvement (EI). SR is defined as 
$SR = \frac{1}{N_{task}}\sum_{i=1}^{N_{task}}R_i$,
where $R_i=1$ if the $i$-th navigation task is successful, otherwise $R_i=0$, and $N_{task}$ is the number of tasks. SPL is defined as
\begin{align}
	SPL = \frac{1}{N_{task}}\sum_{i=1}^{N_{task}}R_i\frac{L_i}{max(D_i, L_i)} \label{eq3}
\end{align}
where $L_i$ denotes the minimum number of steps to find all targets for $N$ agents in the $i$-th task, and $D_i$ is the actual steps for $N$ agents to complete the task. Since $N$ agents navigate to targets simultaneously, $D_i$ is the number of steps of the last agent performing $Done$. EI is calculated as 

\begin{align}
	EI = \frac{1}{N_{suc}}\sum_{i=1}^{N_{task}}B_i\frac{E_i - D_i}{E_i} \label{eq4}
\end{align}
where $B_i=1$ if the $i$-th task is successful with both $N$ agents and a single agent, otherwise $B_i=0$. $N_{suc}$ is the number of tasks that are successful with both the current $N$ agents and a single agent. $E_i$ denotes the number of steps for a single agent to complete the $i$-th task. The value of EI denotes the improved percentage of efficiency of multiple agents compared with an individual agent in the same task.

\subsection{Quantitative Results}

We compare the results of the proposed model with other baselines in unseen testing scenes with both known objects and unknown objects. The \emph{Known Objects} setting means that the target objects in testing scenes have appeared as targets in training tasks, while the \emph{Unknown Objects} setting means that the target objects in testing scenes have not appeared as target objects during training. The quantitative performances in AI2-THOR and RoboTHOR are illustrated in Table \ref{table1} and Table \ref{table2} respectively. \emph{Central}$^{\star}$ denotes that \emph{Central} achieves the highest value in all metrics with different values of $N$, since it can obtain all observations of all agents when making action decisions, which is only regarded as the upper bound for reference. The results of \emph{Central} are marked in italic type, and the best results except for \emph{Central} in each column are marked in bold. We can obtain the following results:

\begin{enumerate}
	
	\item Our model performs better than other baselines except for \emph{Central}. The poor performance of \emph{RGB+IL} and \emph{RGB+SP.+IL}, which are directly extended from single-agent models, indicates that the multi-agent navigation task is challenging and complicated. The high performance of our model demonstrates that our model combining semantic mapping, scene prior knowledge, communication mechanism, and hierarchical decision strategy is reasonable in this multi-agent task.
	
	\item The differences between our model and \emph{Central} in SR, SPL, and EI are the smallest, indicating that the communication mechanism can effectively encode and decode the information and the agent selection strategy of the communication module is reasonable.
	
	\item Models with scene prior knowledge perform better than models of the same structure without scene prior knowledge, which demonstrates that the scene priors can provide effective navigation information to help agents find a shorter path to target objects.
	
	\item Comparing the results on known objects and unknown objects, we can find that our model has better generalization ability compared with other baselines that do not the have semantic mapping module. The information contained in the semantic map and the scene prior knowledge are beneficial for agents to navigate to the target objects that have not been found during training.
	
\end{enumerate}

\subsection{Qualitative Results}

We show some qualitative results in AI2-THOR and RoboTHOR respectively in Fig. \ref{fig:result}. In these samples, the number of agents $N$ is less than the number of target objects $M$, and agents collaborate with others to successfully find all target objects. Agents do not perform $Done$ immediately when finding a certain target object, and they would judge whether they need to continue to navigate to other targets until all target objects are found. Meanwhile, objects that agents see at first may not be close to them, and the object closer to the agent may not be in the current field of view of the agent. Agents can obtain information from others through communication and navigate to the object closer to them with as few steps as possible. The qualitative results of other samples are shown in the attached video.

\subsection{Ablation Experiment}

We conduct the ablation experiment to evaluate the effect of different modules of our model in AI2-THOR. We remove the semantic mapping module to form the model \emph{Ours w/o SemMap.}, and it takes the RGB images as input without building the semantic map. The model \emph{Ours w/o SP.} does not use the scene prior knowledge to build the key objects map. We remove the hierarchical decision module to form the model \emph{Ours w/o HieDec.}, in which the PPO algorithm is trained to directly generate specific actions from $\mathcal{A}$ instead of generating the sub-goals in the high-level decision. 

The results of the ablation experiment indicate that the semantic mapping, scene prior knowledge, and hierarchical decision modules are beneficial for multi-agent navigation. The performances of \emph{Ours w/o SemMap.} decrease the most, indicating that the semantic map plays an important role in exploring and understanding scenes in multi-agent navigation. The results of \emph{Ours w/o SP.} illustrate that scene prior knowledge is beneficial for multiple agents to find shorter paths to navigate to targets. The performance of \emph{Ours w/o HieDec.} demonstrates the effectiveness of the hierarchical decision strategy, and directly predicting specific actions may cause more mistakes in the multi-agent setting.

\section{Conclusion}
In this paper, we propose the multi-agent visual semantic navigation task. We take the advantages of multi-agent exploration to develop a hierarchical decision framework based on semantic mapping, scene prior knowledge, and communication mechanism. The experiment results demonstrate that the proposed multi-agent navigation model can effectively solve this challenging task with a higher success rate and efficiency compared with the single-agent approach.


\bibliographystyle{IEEEtran}
\bibliography{arxiv}

\end{document}